# A Super-Learner with Large Language Models for Medical Emergency Advising

Sergey K. Aityan[1], Abdolreza Mosaddegh[1], Rolando Herrero[2], Haitham Tayyar[3],
Jiang Han[1], Vikram Sawant[1], Qi Chen[1], Rishabh Jain[2], Aruna Senthamaraikannan[2],
Stephen Wood[4], Manuel Mersini[5], Rita Lazzaro[6], Mario Balzanelli[6],
Nicola Iacovazzo[7], Ciro Gargiulo Isacco[8]

[1] Department of Multidisciplinary Engineering, Northeastern University, Oakland, CA, USA
[2] Department of Multidisciplinary Engineering, Northeastern University, Boston, MA, USA
[3] Department of Multidisciplinary Engineering, Northeastern University, Toronto, ON, Canada
[4] Bouvé College of Health Sciences, Northeastern University, Boston, MA, USA
[5] Biovitalage S.R.L., 70010 Valenzano, Italy
[6] Territorial Emergency System SET 118, 74121 Taranto, Italy
[7] Territorial Center for Medical Assistance, 74121 Taranto, Italy
[8] Department of Interdisciplinary Medicine (DIM), Aldo Moro University of Bari, 70121 Bari, Italy

## Abstract

Medical decision-support and advising systems are critical for emergency physicians to quickly and accurately assess patients' conditions and make diagnosis. Artificial Intelligence (AI) has emerged as a transformative force in healthcare in recent years and Large Language Models (LLMs) have been employed in various fields of medical decision-support systems. We studied responses of a group of different LLMs to real cases in emergency medicine. The results of our study on five most renown LLMs showed significant differences in capabilities of Large Language Models for diagnostics acute diseases in medical emergencies with accuracy ranging between 58% and 65%. This accuracy significantly exceeds the reported accuracy of human doctors. We built a super-learner MEDAS (Medical Emergency Diagnostic Advising System) of five major LLMs - Gemini, Llama, Grok, GPT, and Claude). The super-learner produces higher diagnostic accuracy, 70%, even with a quite basic meta-learner. However, at least one of the integrated LLMs in the same super-learner produces 85% correct diagnoses. The super-learner integrates a cluster of LLMs using a meta-learner capable of learning different capabilities of each LLM to leverage diagnostic accuracy of the model by collective capabilities of all LLMs in the cluster. The results of our study showed that aggregated diagnostic accuracy provided by a meta-learning approach exceeds that of any individual LLM, suggesting that the super-learner can take advantage of the combined knowledge of the medical datasets used to train the group of LLMs.

**Keywords**: Super-learner, Meta-learner, Large Language Models, AI-Powered Diagnostics, Multiagent systems, Emergency Medicine

## 1 Introduction

Real-time accurate diagnostics is critically important in emergency medicine because of its time sensitive nature when delays can lead to severe complications or even to patient death. Emergency care practitioners typically work under intense time and resource constraints. The urgency of making a immediate correct diagnosis, inability to conduct extensive observations and collect second opinion, negatively impacts on the ability of making a balanced diagnostic decision (Fleischmann et al, 2016). The elevated level of stress imposed on emergency clinicians adds to the accuracy of diagnostic decisions too (Dias & Neto, 2017; Garcia-Tudela et al, 2022). All above mentioned factors result in burnout syndrome among emergency medical physicians that also reduces their diagnostic accuracy (Boutou, 2019).

As it is reported in American Systematic Review (Newman-Toker et al, 2022), the level of misdiagnoses of traumatic injuries in emergency medicine is quite low ranking about 5%-6% while the rate of misdiagnoses in acute internal emergencies is significantly elevated. For some life threatening and time critical diseases such as stroke, myocardial infarction, aortic aneurysm/dissection, venous thromboembolism, spinal cord compression/injury, and spinal abscess misdiagnosis reaches up to 56% cases. Missed diagnoses for subarachnoid hemorrhage may vary between 0% and 100% for different emergency hospitals. The level of 56% diagnostic errors in spinal epidural abscess was also reported by Bhise et al (2017). The assessment of the level of diagnostic errors in the publications cited above (Newman-Toker et al, 2022; Bhise et al, 2017) was made based on false-positive diagnoses. Also, the misdiagnoses do not include delayed diagnoses. Thus, the number of timely unavailable correct diagnoses may be even higher than reported in the above studies. The initial misdiagnosis and diagnostic delays have been historically described as high as 75%-89% (Tetsuka et al, 2020), which is quite high. Delayed diagnoses in emergency medicine almost equal wrong diagnoses because medical actions to save patients in emergency medicine should be taken without delays.

Diagnostic accuracy is a complement of misdiagnoses to hundred percent. Misdiagnoses may include delayed diagnoses or do not include delayed diagnoses. Such a difference in diagnostic accuracy assessment plays an important role in emergency medicine. Thus, the diagnostic accuracy without accounting for the delayed diagnoses can be assessed as about 40% while the diagnostic accuracy including the delayed diagnoses as failed is in the range of 11%-25% (Tetsuka et al, 2020) which can be averaged up as about 18%. A similar level of 20% diagnostic accuracy by human doctors is reported by King & Nori, (2025).

Quite often, emergency physicians have no time and no practical ability to consult with other medical professionals. Availability of real-time second opinion and practical advice would significantly improve quality of diagnostics and survival rate of emergency patients. Diagnostic errors propagate to clinical decision making, including imaging or test requests, the results interpretation, clinical assessment and illness treatment, particularly in atypical cases or in uncommon pathological conditions (Ronicke et al, 2019).

Thus, the specifics of emergency medicine suggest the need for a medical decision-support diagnostic and advising system that will play a role of a real-time second opinion to enable medical professionals to quickly and accurately assess patients' conditions and make right decisions under stress and time limitations typical in emergency medicine.

Artificial Intelligence has emerged as a transformative force in healthcare, particularly in medical diagnostics. Applying advanced artificial intelligence, including LLMs in emergency medicine provides an opportunity to improve diagnostics, predictive actions, treatment planning, and medication prescription using AI-based decision-support systems. In this paper, we evaluated the performance of multiple top LLMs in medical advising for emergency medicine and introduced a super-learner of LLMs aimed at improving their accuracy and reliability.

## 2 LLMs for Medical Applications

Recent studies have demonstrated the effectiveness of LLMs to aid healthcare professionals in generating comprehensive lists of possible diagnoses based on patient's symptoms, clinical observations, medical history, and other relevant data (Huang et al, 2024; Meng et al, 2024; Nazi & Peng, 2024; Zhou et al, 2025). The utilization of LLMs for medical diagnostics has enhanced accuracy, accelerated decision-making processes, and improved treatment planning. A comparison of physician and AI chatbot responses to public medical questions (Ayers et al, 2023) showed that evaluators preferred chatbot responses (78%) vs responses by human physicians (22%). Singhal et al (2025) reported that Med-PaLM2, a version of PaLM2 fine-tuned on medical data, has achieved a reasonably high accuracy rate in answering medical questions which was close to the accuracy achieved by human clinicians.

The integration of LLMs with medical imaging data is an emerging area of research. Multi-modal models that combine text and image inputs can enhance the interpretation of radiology reports and improve diagnostic accuracy. Brin et al. (2025) assessed performance of a multimodal LLM that integrates visual and textual information to assist radiologists in interpreting medical images. The study shows the potential of zero-shot generative AI in enhancing diagnostic processes in radiology. Huang et al (2024) showed that their LLM chatbot outperformed glaucoma specialists and matched retina specialists in diagnostic and treatment accuracy.

Although pretrained LLMs take advantage of huge training datasets, they may lack specialization in specific domains. Fine-tuning addresses this limitation by allowing the model to learn from domain-specific knowledge to make it more accurate and effective for target application. Emergency medicine has its own unique goals, patterns, and context. Fine-tuning of pre-trained LLMs allows tailoring these models meeting the specific requirements of emergency medicine.

Fine-tuning large LLMs can adapt these pretrained general models on large datasets to specific tasks in medical domain by further training it on clinical data. Yadav et al. (2024) evaluated four LLMs (BART-base, BART-large-CNN, T5 large, and BART-large-xsum-samsum) with and without fine-tuning for diagnostics of mental disorders. The results highlighted that fine tuning significantly improve their performance. and fine-tuned LLMs significantly outperform the same LLMs without fine-tuning.

There are different approaches for fine-tuning of LLMs. Full fine-tuning involves updating all the parameters of a pretrained model on a new task-specific dataset. Devlin et al. (2019) demonstrated this approach demonstrating that the model pre-trained on a large dataset was then fine-tuned on a specific task by adjusting all its weights. This method usually provides a higher level of alignment but can be computationally expensive. On the other hand, quite a few studies focused on parameter-efficient fine-tuning methods that aim to reduce the number of parameters updated during fine-tuning. Rebuffi et al. (2017) proposed fine-tuning only for a small subset of parameters, such as bias terms, which significantly reduced the computational overhead.

Prefix-tuning and prompt-based fine-tuning adjust the input prompts or prefixes fed to the model rather than modifying the model parameters directly. Li & Liang (2021) proposed prefix- tuning, which adds prefixes to the input sequence, allowing the model to adapt to specific tasks. Unlike traditional fine-tuning where the model's parameters are adjusted, prompt-based fine-tuning involves minimal or no changes to the model's weights. Instead, it relies on designing effective queries which result in desired outputs from the model. This method leverages the pretrained model's existing knowledge and can be efficient for LLMs trained in various tasks and knowledgebases.

Some modern LLMs allow fine tuning, however the most advanced commercial LLMs have recently terminated this functionality because they claim to be trained on all available information.

## 3   Evaluating LLMs on Diagnostics in Emergency Medicine

Significant investments of big tech companies in commercial LLMs resulted in models that are generally more powerful and not comparable to their academic or open-source counterparts. However, despite huge investments and technical advantages in LLMs, each of commercial LLMs shows some flaws and limitations in medical diagnosis and treatment specific to each LLM. Fine-tuning reduces these problems but does not eliminate them. Thus, each commercial LLM has its strengths and weaknesses. These shortcomings stem from the complexity of medical diagnostics, limited scope of medical datasets used during pre-training and fine-tuning, which fail to capture the extremely broad variety of diseases, symptoms, test parameters and their correlations, patient conditions, apparent similarity between some diseases, and patient demographic variability.

We have developed a real-time solution for medical emergency diagnostics using commercially available LLMs. For the evaluation purpose, we used GPT, Claude, and Gemini.

A free format inquiry with a case description from an emergency physician was submitted to our system. Automatic prompt generation component processed the inquiry, generated the respective prompts specific to each LLM, and submitted them to respective LLMs. The responses from each LLM were evaluated by a group of medical emergency physicians for diagnosis accuracy, treatment advice, urgency detection, alternative diagnosis, and medical image interpretation using a dataset of 120 real emergency cases. Each response was evaluated using a score from 0 through 4 (0 is performing baseline, 4 is the highest AI performance on current dataset). The final scores were calculated as the average of these scores.

The results of the evaluation of LLM performance on emergency medicine diagnostics are presented in Table 1. As is evident from the table, ChatGPT got the highest score in diagnostic accuracy and was the best but not perfect in treatment advice, and image interpretation, while Claude was the best in urgency detection. All the three LLMs were equally good but not perfect in suggesting alternative diagnoses. On the other hand, Gemini overperforms Claude in image interpretation.

Table 1: Comparing medical capabilities of three LLMs

| Feature | ChatGPT | Gemini | Claude |
|---|---|---|---|
| Diagnostic Accuracy | 4 | 2 | 3 |
| Treatment Advice | 3.2 | 1 | 2 |
| Image Interpretation | 3.3 | 1.7 | 1.3 |
| Urgency detection | 3 | 2 | 3.8 |
| Alternative diagnoses | 3 | 3 | 3 |

The overall better responses received from ChatGPT reflect fine-tuning on emergency cases while Gemini and Claude did not allow fine-tuning and have been used as is.

The results presented in Table 1 do not pretend to represent a comprehensive comparative study of LLMs but used only for the purpose of illustration that different LLMs perform differently on different tasks. We are currently conducting a comprehensive study of variety of commercially available LLMs in medical diagnostics and will report on our findings as soon as this study concludes.

Thus, different LLMs perform differently on different tasks. They also show different performances on diagnostics of different diseases. The results of our comparative study were shown to illustrate that each LLMs has its strength and weaknesses. Thus, if used together as an integrated ensemble, they may provide better results than each LLM alone. The collective capabilities of LLMs in medical diagnosis are expected to be stronger than each individual LLM. This identifies a need for decision support solutions which can take advantage of collective capabilities of LLMs.

## 4    A Super-learner of Large Language Models

We have developed a real-time super-learner solution "MEDAS" (Medical Emergency Decision Advising System) for advising physicians in medical emergency diagnostics. MEDAS stands for Medical Emergency Diagnostics Advising System. The goal of the system is to provide in real time possible diagnoses for an emergency case. The possible diagnoses are generated together with the respective probabilities and urgencies of intervention.

The MEDAS super-learner integrates a number of AI engines which are represented in the current version by several commercially available LLMs such as ChatGPT, Gemini, Claude, Nemotron, and Llama. Any additional AI agent can be easily integrated into the system.

The Super-learner contains a meta-learner which receives responses from different LLMs and learns how to produce optimum results and employ capabilities of each individual LLM (Figure 1).

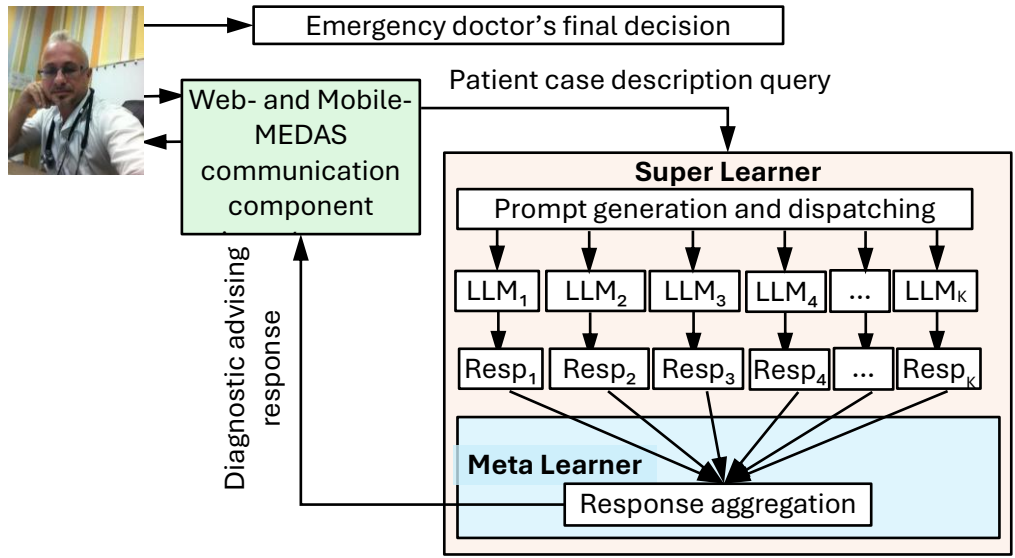

Figure 1: Super-learner of LLMs

In case description inquiry is entered by an emergency physician using the MEDAS communication component via a web browser or the respective app on a smartphone. The inquiry consists of a free format case description that includes all information the physician wants to communicate to the MEDAS advising system. The inquiry can be made as a text or voice entry or a mix of them.

The super-learner prompt generation and dispatching component processes the inquiry and generates the appropriate prompts for each of the integrated LLMs. Then the prompts are sent in parallel to the respective LLMs. Each engaged LLM generates a response to the inquiry that consists of possible diagnoses together with their respective probabilities and urgencies.

All generated responses are submitted to the meta learner component of the super-learner. The meta learner consolidate all responses into one combined response to the emergency physician.

The consolidated response is communicated back to the emergency physician who considers it as second opinion advice. Finally, the emergency physician makes the diagnostic and treatment decision on the case using the MEDAS response to the case.

MEDAS is not making any decisions for the emergency physician but provides all possible diagnoses leaving the final decision on the actual physician.

MEDAS super-learner benefits from all medical datasets used for training a group of LLMs, hence, it can provide more comprehensive information and cover a wider range of health conditions and possible diagnoses. The super-learner also benefits from different capabilities of LLMs in various aspects of emergency medicine.

**Meta-Learner**

Diagnostic capabilities of commercial LLMs are not the same for different diseases. Some LLMs are better or worse in diagnosing some diseases (Aityan et al, 2024). There are also specialized AI agents designed for diagnosing some specific diseases, for example, early diagnostic of sepsis (Aityan et al, 2025). This is due to the differences in their specific focus and differences in medical datasets used for training. The range and content of disorders and symptoms present in those training datasets differ, which can significantly affect the diagnostic capabilities of the LLM. Also, one LLM can provide the best results in diagnostics by symptoms while another LLM might be superior in interpretation of medical conditions. Thus, different LLMs may provide complementary diagnostic results.

The role of a meta-learner is to provide an aggregated response from the super-learner to the physician who asked for diagnostic advice. The meta-learner takes advantage of the best capabilities of each individual LLM. It can learn which LLM is best suited for a specific disease and suggesting a set of more reliable possible diagnoses to optimally combine outputs of the engaged LLMs to achieve the best outcome.

The meta-learner is trained on a dataset containing the responses from the LLMs integrated into the super-learner and confirmed diagnoses provided by medical emergency doctors for patient cases once the diagnoses are finalized. A trained meta-learner can produce optimal results based on collective capabilities of LLMs engaged in the super-learner.

While LLMs are black-box models, the meta-learner model can employ a more comprehensive approach to provide more explicitly understood results. The combination of LLMs as black-box models with the meta-learner provides the super-learner with a certain degree of transparency and explainability valuable for the medical decision-support system.

The meta-learner generates the final response to the diagnostic inquiry by aggregating the results generated by the LLMs integrated into the super-learner. The aggregation mechanism selects the best results generated by the ensemble of LLMs integrated in the supe learner.

In the meta-learner, overall performance in disease diagnostics of each integrated LLM is evaluated, and the respective weights are assigned for the selection of the best responses. These weights are learned through the ongoing training process based on evaluation of the LLMs responses by comparing them with the confirmed diagnoses by medical emergency doctors on real data of medical emergency cases.

For advising on a specific health condition, each individual LLM in the super-learner model receives the same diagnostic inquiry specially configured for each integrated LLM by the automatic prompt generation and generates a response to the received query. Each response from each LLM includes a list (vector) of possible diagnoses, together with the list (vector) of respective probabilities of the suggested diagnoses.

## 5 Super-learner Performance

The performance of the super-learner MEDAS solution was tested with the conventional majority vote used by the meta-learner for each individual LLM integrated in the super-learner, GPT-4o, Claude Opus 4, Llama Maverick 4, Grok 4, and Google Gemini 2.5 Pro.

We tested the accuracy of each of the engaged LLMs on a dataset of 420 real acute internal emergencies cases, which included information on patients' health conditions, medical records, test results, and the corresponding confirmed diagnoses. Then we evaluated the accuracy of the meta-learner aggregated response using the majority vote based on weights adjusted proportionally to the accuracies of the LLMs on the same dataset.

Emergency physicians assessed the responses based on Pass@1 accuracy (the most probable diagnosis) by comparing them to the confirmed diagnoses (Table 2). Only exact matches with the confirmed diagnoses were considered correct while partial alignments were considered incorrect, in accordance with requirements of emergency medicine.

Table 2: Number of correct responses (Pass@1) from 420 emergency cases

| LLMs and Super-learner | Accuracy |
|---|---|
| Gemini | 58% |
| Llama | 59% |
| Grok | 60% |
| GPT | 65% |
| Claude | 65% |
| Super-learner on majority votes | 70% |
| Super-learner, at least one LLM | 85% |

Table 2 shows that accuracy of the individual LLMs vary between 58% and 65% while the meta-learner based on a simple Pass@1 majority vote mechanism responds with the accuracy, 67%, which is higher than any individual LLM alone. However, at least one of the integrated LLMs generates correct response in 85% of cases which is substantially higher. This explicitly demonstrates that a super-learner approach has great potential in advising physicians in emergency diagnostics with much higher accuracy than any individual LLM. A more sophisticated meta-learner model that can learn to choose the best results from the individual LLMs integrated into a super-learner is expected to provide aggregated diagnostic accuracy closer to the above-mentioned 85%. This indicates great potential for the super-learner approach.

To compare the super-learner performance with human physicians, we randomly selected 10 acute internal emergencies cases from the same dataset and asked a group of 12 experienced emergency medicine physicians to provide diagnoses by reading and analyzing acute internal emergenciy cases. This has allowed us to obtain a rough estimate of human doctors' diagnostic accuracy compared to the accuracy of the MEDAS AI solution. Figure 2 shows the results of this experiment with human doctors.

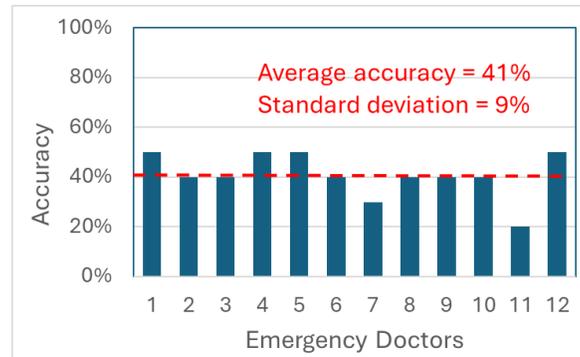

Figure 2: Diagnostic accuracy of 12 human doctors in emergency medicine

The average diagnostic accuracy across all twelve emergency physicians was 41%. With standard deviation 9%. The diagnostic accuracy by human doctors varied between 20% and 50%. These results do not represent a rigorous study but were intended for an estimate of human doctors' diagnostic accuracy. However, our estimates are quite consistent with the diagnostic accuracy by human physicians reported by other research groups (Tetsuka et al, 2020; Newman-Toker et al, 2022; Nori et al, 2025).

Figure 3 shows a comparison of diagnostic accuracy achieved by general medicine doctors (18% on average reported by Tetsuka et al, 2020; 20% reported by King & Nori, 2025), and 43% reported by Newman-Toker et al, 2022), 41% on average by emergency medicine doctors in our own research (Figure 2), individual LLMs, and the super-learner with the same LLMs (Table 2).

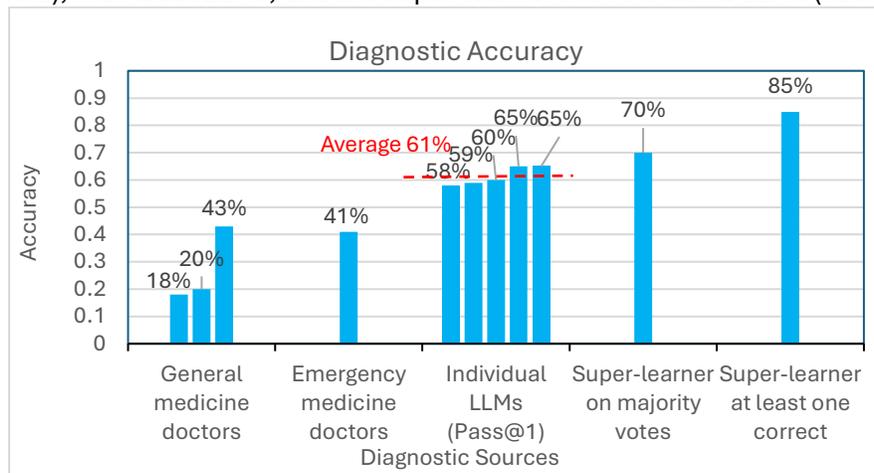

Figure 3: Diagnostic accuracy by general medicine doctors, emergency doctors, individual LLMs, and the super-learner

As is evident from Figure 3, individual LLMs provide higher diagnostic accuracy than human doctors because they have been trained on a much broader volume of clinical information and possess huge amount of knowledge which no individual human is capable to acquire. A super-learner with the same LLMs integrated into a single solution produces higher diagnostic accuracy even with a quite basic meta-learner (~70%). However, at least one of the integrated LLMs in the same super-learner produces 85% correct diagnoses. This clearly suggests that a more sophisticated meta-earner will be capable of significantly improving the diagnostic accuracy of the super-learner by more accurately selecting the best diagnoses generated by the integrated LLMs.

The architecture of the super-learner of commercially available LLMs for medical advising in emergency medicine has evolved from our previous integrated ensemble of AI-based diagnostic solutions including LLMs for emergency medicine (Aityan et al, 2024, 2025a). The super-learner solution may be expanded beyond commercially available LLMs by engaging some other specialized AI-agents in the super-learner solution. For example, we also integrated our proprietary AI-based early sepsis prediction system to the diagnostic advising super-learner as one of the integrated agents (Aityan et al, 2025b).

Microsoft has recently claimed that their AI-based general purpose medical diagnostic solution (King & Nori, 2025) achieves 85% diagnostic accuracy. However, according to their own information, this accuracy is shown as an extreme point on the frontier curve of their performance chart. Such accuracy can be achieved as a target for high-cost diagnostics, which is most likely associated with comprehensive testing and extended time of patient observation. This is not typical for regular emergency medicine. In typical emergency medicine settings, cost, time, and resources for diagnostics in emergencies are normally limited that corresponds to the low-cost range of accuracy of the Microsoft diagnostic frontier which does not exceed 50% or at most 60% as clearly indicated in their performance chart.

# 6  Discussion

We aimed to evaluate the performance of the proposed super-learner approach in real-world emergency medicine settings. To do so, the majority vote, weighted consolidated responses and each individual LLM were assessed separately using the same test dataset.

The results show significant differences in the diagnostic capabilities of LLMs for diagnosis of patients' health conditions in emergency cases.

In this regard, we proposed a super-learner to perform meta-learning over the capabilities of LLMs integrated into the model and train weights based on the performance of the LLMs to improve accuracy. The results highlighted that weighted majority voting can improve the diagnostic outcome. However, the comparison between the accuracy of the aggregated response and that of each individual model shows that the aggregated response exceeds the performance of any individual LLM. This suggests that the super-learner can benefit from the combined knowledge of the medical datasets used to train the engaged LLMs. Thus, it offers more precise advice and covers a wider range of health conditions.

Although the performance of each individual LLM exceeds the performance of human doctors, the collective capabilities of integrated LLMs leveraged by the super-learner can significantly improve overall performance of individual LLMs.

# 7  Limitations and Future Research Directions

In this study, we employed five commercial LLMs for the performance test of the white-box meta-learning approach. Adding more AI agents can leverage the performance of the meta-learning and enrich the results of the study. Also, the weights used in the current study can be fine-tuned to provide better performance.

Our study was primary focused on diagnostics in emergency cases. However, medical advising covers many other aspects including planning treatment and complementary clinical tests. Cooperative multi-agent AI can leverage our super-learner model. Each agent can be a Super-learner itself specialized in one specific task in emergency medicine including diagnosis, treatment, interpretation of medical images and laboratory tests.

Each agent also takes advantage of the expertise of other agents. The agents specialized in diagnostics based on textual or voice notes from doctors may benefit from interpretation of medical images analyzed by other agents and provide inputs for other agents specialized in advising treatments. Such cooperative AI agents enable medical professionals to quickly and accurately assess patients' conditions and make the right decisions under stress and time limitations typical for emergency medicine. Each agent collaborates with other agents using multi-agent coordination techniques based on MCP (Model Context Protocol) design principles which ensure that agents can work together and make joint decisions effectively.

Moreover, some specialized AI models including ad-hoc models for prediction life-threatening diseases and models for analysis patterns from wearables can be added to this agentic framework for leveraging the capabilities of the model.

Our future directions also include work on an efficient self-learning meta-learner which can perform a more comprehensive aggregation of the responses from LLMs and other AI-agents integrated into a super-learner to provide high accuracy combined response to emergency doctors.

The quest for higher diagnostic accuracy may hit fundamental limits of medical diagnostics related to immanent uncertainty caused by complexity of human body. Thus, it is extremely important to investigate and find out such limitations to understand the natural limits of AI approach in medicine.

## 8   Conclusion

Medical diagnostics and treatment in emergency medicine is a complex process characterized by multiple uncertainties, ambiguous symptoms, limited time as well as limited human and clinical resources during emergency. In many cases, even experienced physicians may not arrive at the correct diagnosis based on initial observations. For this reason, a real-time, precise AI-based medical advising system can be lifesaving.

In this regard, we introduced a super-learner to take advantage of collective capabilities of multiple LLMs. The framework can employ both white-box and black-box meta-learner to produce optimal results from outputs of LLMs fine-tuned for emergency medicine.

The results show that the super-learner can take advantage of different capabilities of LLMs in various aspects of emergency medicine since it utilizes the strengths and knowledgebases of multiple LLMs to benefit from their different specializations for diagnostic advising. This approach is applicable to many other decision-support systems dealing with a list of possibilities (e.g., financial decisions) since the accuracy provided by the super-learner model exceeds the accuracy of each individual LLM. The super-learner also provides more reliability for impactful decisions compared to an individual AI agent, since it considers multiple advising responses. Additionally, it avoids hallucinated diagnoses, which are common in LLMs, since it is highly implausible that several LLMs would generate the same hallucinated diagnosis for a single case.

The introduced model takes advantage of the massive investments by big tech companies in commercial LLMs, without needing to build an individual LLM from scratch. It is also dynamically upgradable by integrating new or upgraded LLMs that add capabilities to the system. Moreover, the aggregation mechanism is scalable, as each integrated LLM processes queries independently and in parallel. This makes it possible to add many new LLMs without significant computational cost or aggregation overhead. These characteristics make the super-learner a less expensive yet scalable and more efficient alternative to the huge commercial models for decision support systems. Finally, the successful integration of AI-based medical advising in emergency medicine paves the way for the next generation of autonomous medical agents, which can diagnose diseases and prescribe real-time treatments without human intervention.

## References

Aityan, S. K.; Mosaddegh, A.; Herrero, R.; Inchingolo, F.; Nguyen, K. C.; Balzanelli, M.; Lazzaro, R.; Iacovazzo, N.; Cefalo, A.; Carriero, L.; Mersini, M.; Legramante, J.M.; Minieri, M.; Santacroce,L.; and Gargiulo Isacco, C. (2024). Integrated AI Medical Emergency Diagnostics Advising System. *Electronics*, 13(22), 4389. doi.org/10.3390/electronics13224389

Aityan, S. K.; Mosaddegh, A.; Herrero, R.;  Gargiulo, C.; Lazzaro, R.; and Iacovazzo, N. (2025). *Speech-Driven Medical Emergency Decision-Support Systems in Constrained Environments, in Engineering Cyber-Physical Systems and Critical Infrastructures Networking Data Integrity and Manipulation*, in Cyber-Physical and Communication Systems, Springer, ISSN: 2731-5002, p. 119-140. doi: 10.1007/978-3-031-83149-2_6

Aityan, S.; Herrero, R.; Mosaddegh, A.; Tayyar, H.; Adebesin, E.; Jeedigunta, S.P.; Kim, H. Mersini, M.; Lazzaro, R.; Iacovazzo, N; and Isacco, C.G.;  (2025). AI-Powered Early Detection of Sepsis in Emergency Medicine, Life 15(10): 1576. Doi: 10.3390/life15101576

Ayers, J.W.; Poliak. A.; Dredze, M.; Leas, E.C.; Zhu, Z.; Kelley, J.B.; Faix, D.J.; Goodman, A.M.; Longhurst, C.A.; Hogarth, M.; Smith, D.M. (2023). Comparing physician and artificial intelligence chatbot responses to patient questions posted to a public social media forum. *JAMA Intern. Med*., 183, 589–596. doi:10.1001/jamainternmed.2023.1838

Bhise, V.; Meyer, A.N.D.; Singh, H.; Russo, E.; Al-Mutairi, A; Murphy, D.R.: and Wei, L.  (2017). Errors in Diagnosis of Spinal Epidural Abscesses in the Era of Electronic Health Records, *The American Journal of Medicine*, 130(8), doi: 10.1016/j.amjmed.2017.03.009

Boutou, A; Pitsiou, G.; Sourla, E.; Kioumis, I (2019). Burnout syndrome among emergency medicine physicians: an update on its prevalence and risk factors, European Review for Medical and Pharmacological Sciences, 23: 9058-9065.

Brin, D., Sorin, V.; Barash, Y.; Konen, E.; Glicksberg, B. S.; Nadkarni, G. N.; and Klang, E. (2025). Assessing GPT-4 multimodal performance in radiological image analysis. *European Radiology*, (4):1959-1965. doi: 10.1007/s00330-024-11035-5.

Devlin, J.; Chang, M. W.; Lee, K., & Toutanova, K. (2019). BERT: Pre- training of Deep Bidirectional Transformers for Language Understanding. Proceedings of NAACL-HLT, (1):4171-4186. doi = 10.18653/v1/N19-1423

Dias, R.D. and Neto, A.S. (2017). Acute stress in residents during emergency care: a study of personal and situational factors*, The International Journal on the Biology of Stress*, 20 (3): 241-248. doi.org/10.1080/10253890.2017.1325866

Fleischmann, C.; Scherag, A.; Adhikari, N.K J; Hartog, C.S; Tsaganos, T.; Schlattmann, P.; Angus, D.C; Reinhart, K. (2015). Assessment of global incidence and mortality of hospital-treated sepsis. Current estimates and limitations, *Am J Respir Crit Care Med.*, 2016; 193(3), 259-272.

García-Tudela, Á.; Simonelli-Muñoz, A.J.; Rivera-Caravaca, J.M.; Fortea, M.I.; Simón-Sánchez, L.; González-Moro, M.T.R.; González-Moro, J.M.R.; Jiménez-Rodríguez, D.; Gallego-Gómez, J.I. (2022). Stress in Emergency Healthcare Professionals: The Stress Factors and Manifestations Scale, *Int J Environ Res Public Health*, Apr 5,19(7), 4342. doi: 10.3390/ijerph19074342

Huang, A.S.; Hinabayashi, K.; Barna, L.; Parikh, D.; Pasquale, L.R. (2024). Assessment of a Large language model's Responses to Questions and Cases About Glaucoma and Retina Management, *JAMA Ophtalmology*, 2024 Apr 1;142(4):371-375. doi: 10.1001/jamaophthalmol.2023.6917.

King, D. and & Nori, H. (2025). *The Path to Medical Superintelligence*, Microsoft press release, June 30, 2025, https://microsoft.ai/news/the-path-to-medical-superintelligence/

Li, X. L. and Liang, P. (2021). Prefix-Tuning: Optimizing Continuous Prompts for Generation. *Proceedings of the 59th Annual Meeting of the Association for Computational Linguistics*, (Volume 1: Long Papers), 4582-4597. doi.org/10.48550/arXiv.2101.00190

Meng, X.; Yan, X.; Zhang, K.; Liu, D.; Cui, X.; Yang, Y.; Zhang, M.; Cao, C.; Wang, J.; Wang, X.; Gao, J.; Wang, Y.G.; Ji, JM.; Qiu, Z.; Li, M.; Qian, C.; Guo, T.; Ma, S.; Wang, Z.; Guo, Z.; Lei, Y.; Shao, C.; Wang, W.; Fan, H.; and Tang, YD. (2024). The application of large language models in medicine: A scoping review. *iScience*. 2024 Apr 23;27(5):109713. doi: 10.1016/j.isci.2024.109713. PMID: 38746668; PMCID: PMC11091685.

Nazi, Z. A. and Peng, W. (2024). Large language models in healthcare and medical domain: A review, *Informatics*, Vol. 11, No. 3, p. 57). doi.org/10.3390/informatics11030057

Newman-Toker, D.E.; Peterson, S.M.; Badihian, S.; Hassoon, A.; Nassery, N.; Parizadeh, D.; Wilson L.M.; Jia, Y.; Omron, R.; Tharmarajah, S.; Guerin, L.; Bastani, P.B.; Fracica, E.A.; Kotwal, S.; Robinson, K.A. (2022). Diagnostic Errors in the Emergency Department: A Systematic Review [Internet], *Rockville (MD): Agency for Healthcare Research and Quality (US), Dec. Report No.: 22(23)-EHC043. PMID: 36574484*. doi: 10.23970/AHRQEPCCER258, https://www.ncbi.nlm.nih.gov/books/NBK588123/

Rebuffi, S. A.; Bilen, H.; and Vedaldi, A. (2017). Learning multiple visual domains with residual adapters. *Advances in neural information processing systems*, 30. doi.org/10.48550/arXiv.1705.08045

Ronicke, S.; Hirsch, M. C.; Türk, E.; Larionov, K.; Tientcheu, D.; and Wagner, A. D. (2019). Can a decision support system accelerate rare disease diagnosis? Evaluating the potential impact of Ada DX in a retrospective study. *Orphanet journal of rare diseases*, 14(1): 69. doi: 10.1186/s13023-019-1040-6

Shortliffe, E. H.; Axline, S. G.; Buchanan, B. G.; Cohen, S. N.; and Lederberg, J. (1973). "An artificial intelligence program to advise physicians regarding antimicrobial therapy," *Proceedings of the 3rd International Joint Conference on Artificial Intelligence (IJCAI-73)*, pp. 466–470.

Singhal, K.; Tu, T.; Gottweis, J.; Sayres, R.; Wulczyn, E.; Amin, M.; Hou, L.; Clark, K.; Pfohl, S.R.; Cole-Lewis, H.; Neal, D.; Rashid, Q.M.; Schaekermann, M.; Wang, A.; Dash, D.; Chen, J.H.; Shah, N.H.; Lachgar, S.; Mansfield, P.A.; Prakash, S.; Green, B.; Dominowska, E.; Agüera, Y. A. B.; Tomašev, N.; Liu, Y.; Wong, R.; Semturs, C.; Mahdavi, S.S.; Barral, J.K.; Webster, D.R.; Corrado, G.S.; Matias, Y.; Azizi, S.; Karthikesalingam, A.; Natarajan, V. (2025). Toward expert-level medical question answering with large language models. *Nat Med.*, 31(3):943-950. doi: 10.1038/s41591-024-03423-7. Epub 2025 Jan 8. PMID: 39779926; PMCID: PMC11922739.

Tetsuka, S.; Suzuki, T.; Tomoko Ogawa, T.; Hashimoto, R.; and Kato, H. (2020). Spinal Epidural Abscess: A Review Highlighting Early Diagnosis and Management, JMA J. 2020 Jan 15;3(1):29-40. doi: 10.31662/jmaj.2019-0038.

Yadav, M.; Sahu, N. K.; Chaturvedi, M.; Gupta, S.; and Lone, H. R. (2024). *Fine-tuning Large Language Models for Automated Diagnostic Screening Summaries*. arXiv preprint arXiv:2403.20145. doi.org/10.48550/arXiv.2403.20145.

Zhou, Z.; Xu, Z.; Zhang, M.; Xu, C.; Guo, Y.; Zhan, Z.; Fang, Y.; Dong, S.; Wang, J.; Xu, K.; Xia, L.; Yeung, J.; Zha, D.; Cal, D.; Melton, G.B.; Lim, M. and Zhang, R. (2025). Large language models for disease diagnostics; a scoping review, *Cornel Univdersity, arXiv:2409.00097*, https://arxiv.org/pdf/2409.00097